\documentclass{article}
\usepackage{ijcai21}

\usepackage{times}
\usepackage{soul}
\usepackage{url}
\usepackage[hidelinks]{hyperref}
\usepackage[utf8]{inputenc}
\usepackage[small]{caption}
\usepackage{graphicx}
\usepackage{amsmath}
\usepackage{caption}
\usepackage{subcaption}
\usepackage{amsthm}
\usepackage{comment}
\usepackage{cases}
\usepackage{mathtools}

\usepackage{amsmath,amssymb}
\DeclareMathOperator{\E}{\mathbb{E}}
\usepackage{booktabs}
\usepackage{algorithm}
\usepackage{algorithmic}
\makeatletter

\newcommand*\titleheader[1]{\gdef\@titleheader{#1}}

\AtBeginDocument{%

\let\st@red@title\@title

\def\@title{%

\bgroup\normalfont\large\centering\@titleheader\par\egroup

\vskip1.5em\st@red@title}

}

\makeatother

\title{Deep hierarchical reinforcement agents for automated penetration testing}
\titleheader{\emph{IJCAI-21 1\textsuperscript{st} International Workshop on Adaptive Cyber Defense}}

  \author{
  Khuong Tran$^1$
  \and
  Ashlesha Akella$^1$\and
  Maxwell Standen$^2$\and
  Junae Kim$^2$\and 
  David Bowman$^2$\and
  Toby Richer$^2$\and 
  Chin-Teng Lin$^1$
  \affiliations
  $^1$University of Technology Sydney, Australia\\
  $^2$Defence Science and Technology Group, Australia\\

  \emails
  \{Khuong.Tran $|$ Ashlesha.Akella $|$ Chin-Teng.Lin\}@uts.edu.au,
  \{Max.Standen $|$ Junae.Kim $|$ David.Bowman $|$ Toby.Richer\}@dst.defence.gov.au
  }

\begin{document}

\maketitle

\begin{abstract}
  Penetration testing – the organised attack of a computer system in order to test existing defences – has been used extensively to evaluate network security. This is a time consuming process and requires in-depth knowledge for the establishment of a strategy that resembles a real cyber-attack. This paper presents a novel deep reinforcement learning architecture with hierarchically structured agents called HA-DRL, which employs an algebraic action decomposition strategy to address the large discrete action space of an autonomous penetration testing simulator where the number of actions is exponentially increased with the complexity of the designed cybersecurity network. The proposed architecture is  shown to  find  the  optimal attacking policy  faster  and  more  stably than a conventional deep Q-learning agent which is commonly used as a method to apply artificial intelligence in automatic penetration testing.

\end{abstract}

\section{Introduction}
The development of effective autonomous defenses requires sophisticated attacks to train against. Penetration Testing (PT) has been used extensively to evaluate the security of ICT systems. This is a time-consuming process and requires in-depth knowledge for the establishment of a strategy that resembles a real cyber-attack. Available automatic penetration testing tools cannot learn to develop new strategies in response to simple defensive measures. As a result, automatic penetration testing tools that use artificial intelligence are desirable for simulating real attackers with an attacking strategy instead of testing all possible values when hacking into a system.

Reinforcement Learning (RL) is a paradigm of learning in which an autonomous decision-making agent explores and exploits the environmental dynamics and discovers a suitable policy for acting in such environments. This makes RL a suitable candidate for tackling the problem of automating the PT process. Deep reinforcement learning (DRL) with function approximators as neural networks have achieved many state of the art results on a variety of platforms and applications, as demonstrated by ~\cite{mnih2015human} and ~\cite{lillicrap2015continuous}. Value-based methods such as variants of deep Q-learning (DQN) algorithms ~\cite{watkins1992q} have become the standard in many approaches thanks to their simplicity and effectiveness in learning from experienced trajectory. However, one of the limitations of applying DQN in practical applications is the complexity of action space in different problem domains. DQN requires an evaluation for all actions at the output and uses a maximization operation on the entire action space to select the best action to take in every step. This can be problematic if the action space is large as multiple actions may share similar values ~\cite{DBLP:journals/corr/abs-1906-12266}. PT is one such practical application where the action space scales exponentially with the number of host computers in different sub-networks. Applying conventional DRL to automate PT would be difficult and unstable as the action space can explode to thousands even for relatively small scenarios ~\cite{schwartz2019autonomous}.    

There are a number of approaches to address the intractability problem of a large discrete action space, such as work done by ~\cite{tavakoli2018action} and ~\cite{dulac2015deep} where the action space is hierarchically structured into branches or prior domain knowledge is applied to embed the action space respectively. This study aims at introducing a new action decomposition scheme which is efficient and versatile in developing PT attacking strategy in different network configurations. The proposed method is inspired by the paradigm of multi-agent RL to decompose the action space into smaller sets and train individual RL agents in each of the action subsets. 

We begin by giving a brief overview of the available literature on the problem, followed by the background and notations. A detailed description of our proposed method is then provided together with empirical results which demonstrate the performance of our algorithm. We use CybORG which is an autonomous cybersecurity simulator platform developed by ~\cite{baillie2020cyborg} to empirically validate the learning and convergence properties of our architecture.

\section{Related Work}
RL has been used recently as an AI method to model and develop exploit attack to cybersecurity setting as done ~\cite{schwartz2019autonomous}, ~\cite{hu2020automated} and ~\cite{zennaro2020modeling}. In these works, PT is formulated as a partially observable markov decision process (POMDP) where RL or DRL is used to model the environment and learn exploit strategy. The attacker agent only observes the compromised hosts and/or subnets but not the entire network connections. The agent then needs to learn from trial-and-error experiences in order to develop suitable policy or strategy to gain access to hidden assets. 
These previous studies have shown tabular RL or DRL to be a possible method to solve small scale networks or capture the flag challenges which has relatively small action spaces. Extending DRL to networks with a larger action space is still a standing challenge not only in cybersecurity setting but also in RL context ~\cite{schwartz2019autonomous}.       

~\cite{dulac2015deep} proposed an architecture, named Wolpertinger, that leverages the actor-critic framework of RL. This work attempted to learn a policy in a large action space problem with sub-linear complexity. It uses prior domain knowledge to embed the action space and applies nearest-neighbor method to cluster similar actions to narrow down the most probable action selection. This method can be unstable in a sparse reward environment as  the gradient cannot propagate back to enhance the learning of the actor network. 

Another study done by ~\cite{DBLP:journals/corr/abs-1809-02121} suggested the use of an elimination signal to teach the agents to ignore irrelevant actions to reduce the number of possible actions from which to choose. A rule-based system calculates the action elimination signal by evaluating actions and assigning negative points to irrelevant actions. Then, the signal is sent to the agent as part of an auxiliary reward signal to help the agent learn better. This framework is best suited to domains where a rule-based component can easily be built to embed expert knowledge and facilitate the agent's learning by reducing the size of the action set.

~\cite{tavakoli2018action} incorporates different network branches to handle different action dimensions. A growing action space is trained using curriculum learning to avoid overwhelming the agents  ~\cite{DBLP:journals/corr/abs-1906-12266}. ~\cite{DBLP:journals/corr/abs-1902-00183} focused on learning action representation so the agent can infer similar actions using a single representation while work done by ~\cite{vandewiele2020qlearning}  targeted at learning action distributions. All these approaches aimed at finding a way to reduce the large number of actions. However they cannot be applied to solve automating PT for the following reasons. Firstly, the action space in PT is entirely discrete without continuous parameters. This is different with other works where the action space is parameterised and comprised of few major actions within which contain different continuous parameters. Secondly, each action in PT can have very different effects such as attacking hosts in different subnets or different method of exploits. This nature is again different with large discretised action space, where group of actions can have spatial or temporal correlation , which is addressed in ~\cite{tavakoli2018action}, ~\cite{vandewiele2020qlearning} or ~\cite{dulac2015deep}.

To the authors' best knowledge, this work is the first to propose a non-conventional DRL architecture to directly target the large action space problem in PT settings. An explicit action decomposition scheme using an hierarchical agents reinforcement learning approach (where the agents are grouped hierarchically and each agent is trained to learn within its action subset using the same external reward signal) is used. This proposed model-free methodology requires minimum domain knowledge to decompose the set of actions and is proven to be efficient in scaling for more complex problems. The solution also takes advantage of value based reinforcement learning to reduce the sample complexity and it can be modified to extend to other problem domains in which the action space can be flattened into a large discrete set. 

\section{Problem description}
The problem to be considered in this paper is a discrete time reinforcement learning task modelled by an MDP \footnote{We do not want to formulate PT as POMDP similar to other works as to clearly demonstrate the effect of the proposed architecture without compounding factor, our method can still be applied as a feature to POMDP solution. Additionally, none of previous works uses any POMDP specific solutions}, formalised by a tuple $\langle \mathcal{S}, \mathcal{A}, \mathcal{P}, \mathcal{R}, \mathcal{\gamma}  \rangle$. At each time step $t$, the agent receives a state observation $s_t \in \mathcal{S}$ from the environment. In our PT setting, this represents the observable part of the compromised network. To interact with the environment, the agent executes an action $a_t \in \mathcal{A}$, after which it receives another state $s_{t+1}$ according to the environment transition function $\mathcal{P}: \mathcal{S} \times \mathcal{A} \times \mathcal{S} \rightarrow \mathbb{R}$, and a reward signal $r_t$ given by $\mathcal{R}: \mathcal{S} \times \mathcal{A} \rightarrow \mathbb{R}$. The agent aims to maximise a utility, defined as the total sum of all reward gained in an episode $R_t = \sum_{\tau=t}^{\infty}\gamma^{\tau-t}r_{\tau}$, where $\gamma \in [0,1]$ is the discount factor used to determine the importance of long term rewards. 

Value-based methods such as DQN algorithms learn a (near) optimal policy $\pi: \mathcal{S} \rightarrow \mathcal{A}$ which maps the observed states or observations to actions. The action value function as defined below tells each agent $i$ the expected utility of choosing the action $a^i_t$ given state $s_t$, and the expected utility for that state:
    \begin{equation}
        \label{eqn:action_value_function_general}
        \begin{aligned}
        Q^{\pi,i}( s,a^i ) & = \E[ R_t | s_t=s, a_t=a^i,\pi^i ] \\
        & = \E_{s'}[r + \gamma\E_{a^{i'} \sim \pi^i(s')}[Q^{\pi,i}(s',a^{i'})] | s,a^i,\pi^i]
        \end{aligned}
    \end{equation}
    

By recursively solving Equation \ref{eqn:action_value_function_general}, the optimal action value function defined as $Q^*(s,a) = \max_{\pi} Q_{\pi}(s,a) $ can be attained by taking the action with highest Q-value in the next state. This solution is shown in Equation \ref{eqn:action_value_function_optimal} and this is the main equation of the popular Deep Q-Learning (DQN) algorithm.
    \begin{equation}
        \label{eqn:action_value_function_optimal}
        \begin{aligned}
        Q^{i*}(s,a^{i}) = \E_{s'}[r + \gamma \max_{a^{i'}}Q^{i*}(s',a^{i'}) | s,a^i]
        \end{aligned}
    \end{equation}

In non-trivial problems, Q values are often approximated by parameterised functions such as neural networks (e.g. $Q(s,a; \theta)$). 
If the action space $|\mathcal{A}|$ is large, DQN becomes unstable as there are many different actions with similar Q-values and it is not clear which action should be taken. As a result, the performance of DQN and its variants degrade as the action space increases ~\cite{dulac2015deep}. The action space in reinforcement learning has different formats and representation depending on the type of problems. Mainly they are classified into two main categories: continuous and discrete. In a discrete action problem, the action space can be organised to have a flat action space or parameterised into an hierarchically structured action space where there are a few types of main actions under which there are sub-actions \cite{DBLP:journals/corr/MassonK15}. It is harder, however, to learn in a parameterized action space as the agent needs to learn two Q-functions or policies, one for choosing the main action and another to choose the sub-action or the parameters of each action. On the other hand, converting the hierarchically structured action space into a flat representation could result in a substantially larger action size which also hinders the learning of DQN agents. This research proposes a multi-agent learning framework where the agents are grouped into an hierarchical structure to tackle this particular instance of a large flat discrete action space, or a flattened parameterised action space.

\begin{figure}[ht]
    \centering
    \includegraphics*[scale=0.3]{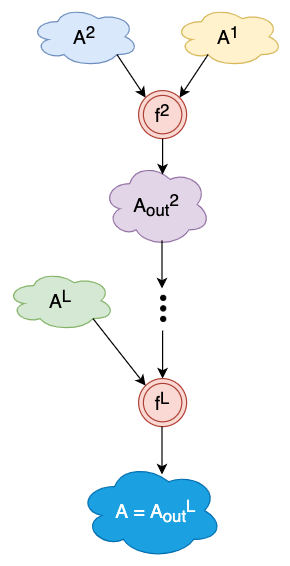}
    \caption{Action space composition}
    \label{fig:action_space}
\end{figure}

In a flat and discrete setting, the action space is encoded into integer identifiers ranging from $LOW\_INT$ to $MAX\_INT$, for instance from $0$ to $999$ in an action space of $1000$ actions. Each number is translated into a valid action that can be understood and executed by the simulator. In our work, we do not assume any prior knowledge over the semantics or relationships among these individual actions. The main idea of the proposed solution is to decompose the overall action space $\mathcal{A}$ into smaller sets $\mathcal{A}^1, \mathcal{A}^2, \dots, \mathcal{A}^L$, where $|\mathcal{A}|^{i} << |\mathcal{A}|$. If these sets are combined together using pre-defined linear functions (e.g $f^{i+1} : \mathcal{A}^{i} \times \mathcal{A}^{i+1} \rightarrow \mathcal{A}_{out}^{i+1}$), they can produce a strictly increasing sets of action representation: $|\mathcal{A}_{out}^1| < |\mathcal{A}_{out}^2| < \dots < ... < |\mathcal{A}_{out}^L|$ with $\mathcal{A}_{out}^L = \mathcal{A}$. Each $\mathcal{A}_{out}^{i}$ is a set of action representation constructed up to level $i$. The sets $\mathcal{A}^{i}$ and $\mathcal{A}^{i+1}$ can be considered as the supports of the set $\mathcal{A}_{out}^{i+1}$. Intuitively, the integer identifiers of the final actions which can be large in value are built up algebraically using smaller integer values.  
The growing numerical representation of the action space is illustrated in Figure \ref{fig:action_space}.

\section {Hierarchical-Agent Action Decomposition}
\subsection{Action decomposition scheme}

The main contribution of this paper is to propose an action decomposition strategy to reduce the large discrete action space into manageable sets for deep reinforcement learning agents, which stabilises the learning and enhances convergence properties.   

\begin{figure}[ht]
\centering
\includegraphics*[scale=0.3]{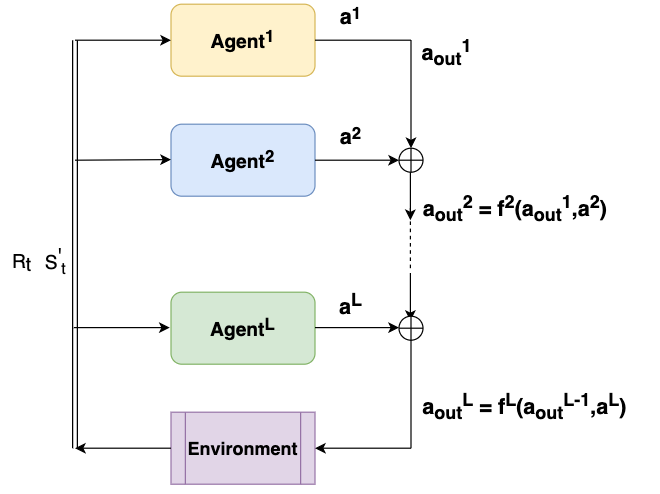}
    \caption{HA-DRL Architecture}
    \label{fig:model}
\end{figure}

As mentioned in the previous section, the large action space is decomposed into $L$ sub-action spaces. Instead of using different neural networks' heads to manage these action subsets as proposed by ~\cite{tavakoli2018action}, separate DQN agents are used to control each of these subsets. The benefits of this approach will be highlighted in the next section.
The agents at levels $i$ and $i+1$ output \textit{primitive} action signals $a^{i}$ and $a^{i+1}$, each of which is within the ranges of $[0, \dots,|\mathcal{A}^i| - 1]$ and $[0, \dots,|\mathcal{A}^{i+1}| - 1]$ respectively. These primitive actions do not interact directly with the environment but they are used to build up the final action. Each agent from level $1$ to level $L$ can use its action signal, together with the external reward $R_t$, to train its own neural network using DQN updates as in Equation \ref{eqn:action_value_function_optimal}. 
The primitive action values are then combined together using a linear function to produce the action representation at level $i+1$: $a_{out_{t}}^{i+1} = f^{i+1}(a_{out_{t}}^{i}, a_t^{i+1}) = a_{out_{t}}^{i} \times \beta^{i+1} + a_t^{i+1}$, where $\beta^{i+1}$ is a pre-defined value used to scale up the output of $f^{i+1}(a_{out_{t}}^{i}, a_t^{i+1})$ and $\beta^{i+1} =  |\mathcal{A}^{i+1}|$. To simplify notations, time step $t$ will be removed to enhance equations readability, therefore: $a_{out}^{i+1} = f^{i+1}(a_{out}^{i}, a^{i+1})$ with $a_{out}^1 = a^{1}$. For instance, we can decompose an action space of a given scenario with $|\mathcal{A}| = |\mathcal{A}_{out}^{L}| = 1000$, using logarithm of the original action set with base $10$, into $L$ sets where: $L = \log_{10}{|\mathcal{A}_{out}^{L}|} = 3$. $\mathcal{A}^1$, $\mathcal{A}^2$ and $\mathcal{A}^3$ all contain 10 action identifiers from $0$ to $9$. When we combine together $\mathcal{A}^1$ and $\mathcal{A}^2$ using the function $f^2(a_{out}^{1}, a^{2}) = a_{out}^{1} \times \beta^{2} + a^{2}$ with $\beta^{2} = 10$, $\mathcal{A}_{out}^2$ will expand the range of actions it can represent to be from $0$ to $99$ which has $100$ actions in total. Finally to arrive at the action space of $1000$ actions, we apply the same procedure to $\mathcal{A}^3$ with $a_{out}^3 = a_{out}^{2} \times \beta^{3} + a^{3}$ and $\beta^{3} = 10$. As a result, $\mathcal{A}_{out}^3$ contains action values from $0$ to $999$. In summary, after a few levels of function compositions, the agent uses the final action, which is computed via $a_{out}^{L} = f^{L}(a_{out}^{L-1}, a^{L})$, to interact with the environment. The working mechanism of the proposed framework is depicted in Figure \ref{fig:model}. We name this method as Hierarchical Agent Deep Reinforcement Learning or HA-DRL.


\subsection{Hierarchical Agents Deep Reinforcement Learning (HA-DRL)}

HA-DRL extends the conventional MDP framework with a single reinforcement learning agent to having a factored action space handled by a group of DQN agents. This means that the original action value function $Q(s, a; \theta)$ is decomposed into a combination of smaller action value functions $Q(s, a^{i}; \theta^{i})$ where $i$ is in the range of $1,\dots, L$. 
This approach has three major advantages in facilitating learning and enhancing the convergence property.

First of all, leveraging a system of hierarchical agents in learning action subsets instead of having a single large DQN agent with thousands of actions in the output improves the use of the the learning signal to update different subsets of $\theta$. Each agent uses the same reward signal to minimise its own loss function as in Equation \ref{eqn:loss_agent_function} and \ref{eqn:target}. The parameterised policy $\pi^{i}(a^{i}|s; \theta^{i})$ is updated to produce optimal abstract actions. As long as each agent learns the optimal policy to achieve highest possible long term reward, together with substantial exploration, the combined action or the overall policy will converge towards optimal behavior. This reasoning is formalised as follows:
\begin{equation}
        \label{eqn:optimal_action_value_func}
        \begin{aligned}
        Q^*(s_t, \boldsymbol{a}; \theta) & = \E_{\pi}[ r_t+ \gamma\max_{\boldsymbol{a}^{'}}Q^*(s_{t+1}, \boldsymbol{a}^{'}) | s_t = s, a_t = \boldsymbol{a}] 
        \end{aligned}
    \end{equation}

with $\boldsymbol{a} = f(a^1, a^2, \dots, a^L)$.

The above optimal state-action value function is achieved when the agents can maximise the future expected environment reward, which is the same goal for each of the agent in the hierarchy. With enough exploration, the group of agents can explore all possible combinations of individual actions to achieve this global optimal policy. However with sparse reward scenarios or extremely large action space, this condition may not be practically feasible. Current empirical results in tested scenarios showed this approach to result in faster training and better convergence properties than a single DQN with combined neural network.

Secondly, the neural networks of HA-DRL can be constructed with a smaller neural network for each of the agents. HA-DRL does not solve a complex problem by multiplicatively duplicating a single DQN agent into multiple agents. It does so by breaking down the action space into small sets so that each agent does not have to learn all the intricate features of the state to acquire the overall optimal policy. As a result, each agent in HA-DRL needs a smaller neural network which is enough to map relatively abstract state features to action representations. Additionally the required number of agents is not large even if we have thousands of actions as we will discuss in the following section. This justifies the concerns over network complexity of having multi-agent learning.

\begin{figure}[ht]
\centering
\includegraphics*[scale=0.45]{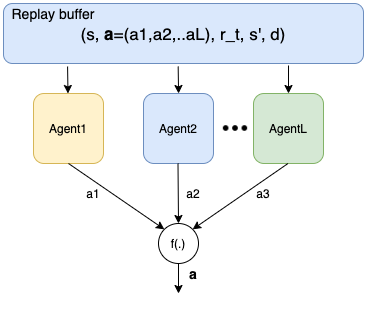}
    \caption{Parallel training and execution}
    \label{fig:parallel_training}
\end{figure}

Lastly, the training and inference of all agents do not have to be sequential in nature. Only the application of the linear function composition has to be constructed sequentially which is almost a constant operation with respect to the increase in complexity of the action space. With appropriate configurations of multi-agent DQN training and the use of parallel programming frameworks such as Pytorch, all agents can be trained and used for inference in parallel which produces no extra cost compared to a single DQN training. This is also true for the utilisation of a replay buffer where a single replay buffer is used for all agents since the tuple of information is mostly the same for all agents, except for the outputted action of each agent (Figure \ref{fig:parallel_training}). 

The learning of all agents in this paper follows the Dueling-DQN (DDQN) architecture as introduced by ~\cite{wang2015dueling} where $\theta^{i}$ is updated by optimizing the loss: 

\begin{equation}
        \label{eqn:loss_agent_function}
        \begin{aligned}
        L^i(\theta^i) = \E_{s_t,a_t^{i},r_t,s_{t}'}[(y^{i,DDQN} - Q^i(s_t, a_{t}^{i}; \theta^{i})^2]
        \end{aligned}
    \end{equation}

with 

\begin{equation}
    \label{eqn:target}
    \begin{aligned}
    y^{i, DDQN} = r_{t} + \gamma \max_{a'}Q(s', {{a}^{'}}^{i}; {{\theta}^{-}}^{i})
    \end{aligned}
\end{equation}

where ${{\theta}^{-}}^{i}$ is the parameter of the \emph{target network} which is mentioned in the original DQN paper \cite{mnih2015human}.  

\subsection{Interpretability of action decomposition}

\begin{figure}[ht]
\centering
\includegraphics*[scale=0.3]{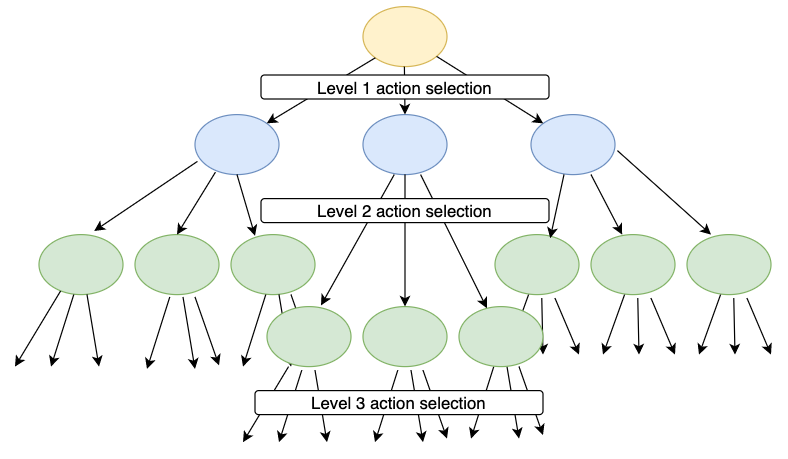}
    \caption{Hierarchical Action Selection}
    \label{fig:hierarchical_action}
\end{figure}

The idea of decomposing a large action space into smaller sets is similar to having a tree structure of multiple layers of actions. Each node in the tree acts like an action selection node where it picks the corresponding action from its child nodes. Each action node has its own range of integer identifiers. In order to reach a leaf node with a specific action value, the parent node needs to pick the right children to reach a specified leaf. The Figure \ref{fig:hierarchical_action}, which is an adaptation from a structured parameterised action space in \cite{DBLP:journals/corr/abs-1903-01344}, illustrates the idea of having different layers of action selection.

Instead of having a different number of agents in each level to represent ranges of integer values, we leverage the linear algebraic function to shift the values of action identifiers into appropriate ranges. The function $a_{out}^{i+1} = f(a_{out}^{i}, a^{i+1}) = a_{out}^{i} \times \beta^{i+1} + a^{i+1}$ where the coefficients $\beta^{i+1}$ represent the number of action selection nodes or agents at level $i+1$. The value of $\beta^{i+1}$ shifts the output range of the integer values of the actions at a particular level. This results in having only one agent per level as shown in Figure \ref{fig:model}.
This action decomposition scheme shows that instead of having a single RL agent explore and learn an entire action space of $|\mathcal{A}|$ which can be extremely large, a few smaller RL agents can be used to explore and learn on a much smaller version of the action space of its own. The outputs of these smaller agents are then chained together to produce the final action.  
The complexity of having multi-agents on different levels is similar to traversing a tree in computer science. The number of levels $L$ is in the order of $L \approx \log |\mathcal{A}|$. As a result, it requires only a handful of agents for an action space even with millions of actions.

\subsection{Optimality of action value function}
This subsection discusses the optimality of the learned policy under this action decomposition scheme. The optimal action value function $Q^{*}(s, a; \theta)$ is approximated by the optimal action value function of each agent $i$ with ${Q^{*}}^{i}(s, a^{i}; \theta^{i})$. As the final action is actually the combination of all previous action values
$a = f(a^1, a^2,\dots, a^L)$, the optimal action value function is related to the combination of individual optimal action value function via a function $g(.)$:

    \begin{equation}
        \label{eqn:new_action_value_func}
        \begin{aligned}
        Q^*(s,a; \theta) & = Q^*(s, f(a^1, \dots , a^L); \theta^1,\dots, \theta^{L})  \\
            & = g({Q^{*}}^{i}(s, a^i; \theta^{i}))
        \end{aligned}
    \end{equation}

As the action of one agent can have an effect on the reward received by another agent, the global optimal policy is not guaranteed in complex scenarios or with limited and finite exploration. Current works are being done to implement coordinator module such as QMIX proposed by \cite{DBLP:journals/corr/abs-1803-11485} to approximate the function $g(.)$. 

However for medium-sized scenarios with frequent reward signals, the proposed architecture can handle the learning better than individual DRL agent.

\section{Experiment Setup}

\subsection{CybORG simulator}
We use an autonomous penetration testing simulator as our testbed for validating the performance of HA-DRL. ~\cite{baillie2020cyborg} developed CybORG to provide an experimental environment for conducting AI research in a cybersecurity context and it is designed to enable an autonomous agent to conduct a penetration test against a network. Apart from being open source and having the appropriate OpenAI gym interface for applying different DRL algorithms, this environment represents a real application where we are faced with different aspects of complexity: sparsity of reward signals, large discrete action space and discrete state representation mean we cannot rely on any computer vision techniques to facilitate the learning. We summarize the details of the execution and complexity of CybORG here.  

The simulation provides an interface for the agent to interact with the network and provides a vector to the agent that is based on the observed state of the network. Through this interface, the agent performs actions that reveal information about the simulated network and successfully penetrates the network by capturing the flags on the hosts within it.

\begin{figure}[H]
 \centering
      \includegraphics*[scale=0.3]{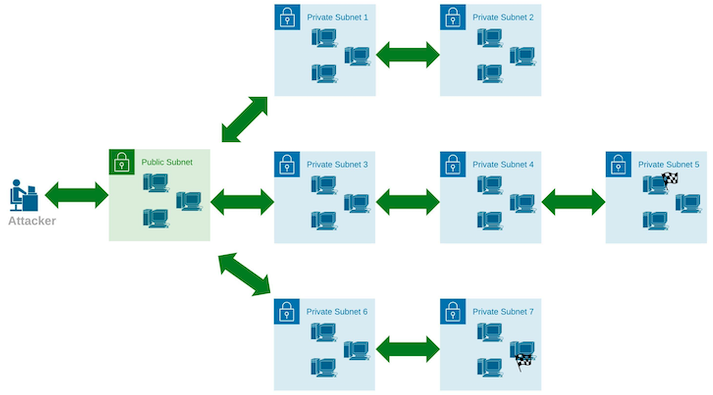}
      \caption{An example of 24 hosts scenario with $1$ public subset and $7$ private subnets where flags represent valuable assets are hidden in private subnet $5$ and $7$}
      \label{fig:simulator}
\end{figure}

The simulation can implement various scenarios. These scenarios detail the topology of the target network, the locations of the flags, and host specific information, such as the operating system and running services. The target network features a series of subnets that contain hosts. The hosts in a subnet may only act upon hosts in the same subnet or adjacent subnets. An illustrative representation of a possible scenario configuration with 24 hosts and one attacker agent is presented in Figure \ref{fig:simulator}. In this example scenario, there are two hidden flags in one of the hosts in subnet 5 and in one of the hosts in subnet 7. There are 3 hosts in each of the subnets. The scenarios differ in terms of the number of hidden flags and the locations of the hosts that contain the flags, along with the number of hosts, which determines the complexity of the state and action spaces. 

In the 24-host example, 550 actions are available to the attacker agent. The simulation includes three action types: host-to-host actions, host-to-subnet actions, and on-host actions. Each action may be performed on a host, and each host-to-host action or host-to-subnet action may target another host or subnet, respectively. These actions include the following:
\begin{itemize}

    \item actions that reveal information about a host’s operating system (OS),
    \item actions that reveal network information that is contained on a host,
    \item actions that reveal information about other hosts in the network using scans with an internet control message protocol (ICMP) ping,

    \item actions that reveal information about the services on a host with a transport control protocol (TCP) SYN sweep,

    \item actions that provide information about the network that is obtained via passive observation of traffic,
    \item actions that provide remote code execution (RCE) from the exploitation of a vulnerable service with brute-force guessing of SSH credentials. The agent can use RCE on a host to pivot to other hosts.
    
\end{itemize}

For a specified scenario, $p$ hosts are grouped into $q$ subnets. When the simulation supports $m$ types of host-to-host actions, $n$ types of host-to-subnet actions and $o$ types of of host actions, the action space for host-to-host actions is $m \times p \times (p-1)$, the action space for host-to-subnet actions is $n \times p \times q$, and the action space for on-host actions is $o \times p$. This leads to a large potential action space even in a small network.

The success of actions depends both on the state of the simulation and the probability. The state of the simulation consists of various information, such as the operating system on each host and the services that are listening on each host. The agent sees an unknown value for each value it does not know. As the agent takes actions in the simulation, its awareness of the true state of the simulation increases. An action fails if the agent does not have remote access to a target host or subnet. However, an action also has a probability of failing even if the agent has access to the target. This probability represents that attempts to exploit vulnerabilities are seldom if ever 100\% reliable.

The simulation was deliberately configured to pose a highly sparse environment. The agent began the game with no knowledge of the flag location so finding the host with the flag or a hidden asset was completely random. A final high reward is given for capturing the flag while a small positive reward is given for every host that was successfully attacked. The flag was hidden from the state space; hence, the agent could not determine the location of the flag by observing the state vector.
This simulation differs significantly from the typical RL environments, in which either the objective or target is visible to the agent.

\subsection{Neural network architecture}

Each agent in our experiments uses a 3-layer Dueling DQN network architecture. However HA-DRL can be used with any variants of DQN algorithms such as prioritised experience replay, double DQN etc..
The performance of HA-DRL is compared against a single Dueling DQN agent.
The first two layers have $2048$ neurons to evaluate the advantage functions while the last layer has $512$ neurons to compute the value functions. The action space of individual DQN agents in HA-DRL is limited to be in the range of $8$ to $10$. As a result, there are at most 4 DQN agents created for scenarios of up to $5000$ actions. Due to the complexity of CybORG simulator, the algorithm's performance was demonstrated on a maximum complexity of $100$ hosts which has the action space of $4646$ actions. All the training runs are conducted using a single RTX6000 GPU. 
The implementation can be shared upon requests from interested readers.

\section{Results}

\subsection{Policy training}
\begin{figure}[ht]
    \centering
    \includegraphics[scale=0.19]{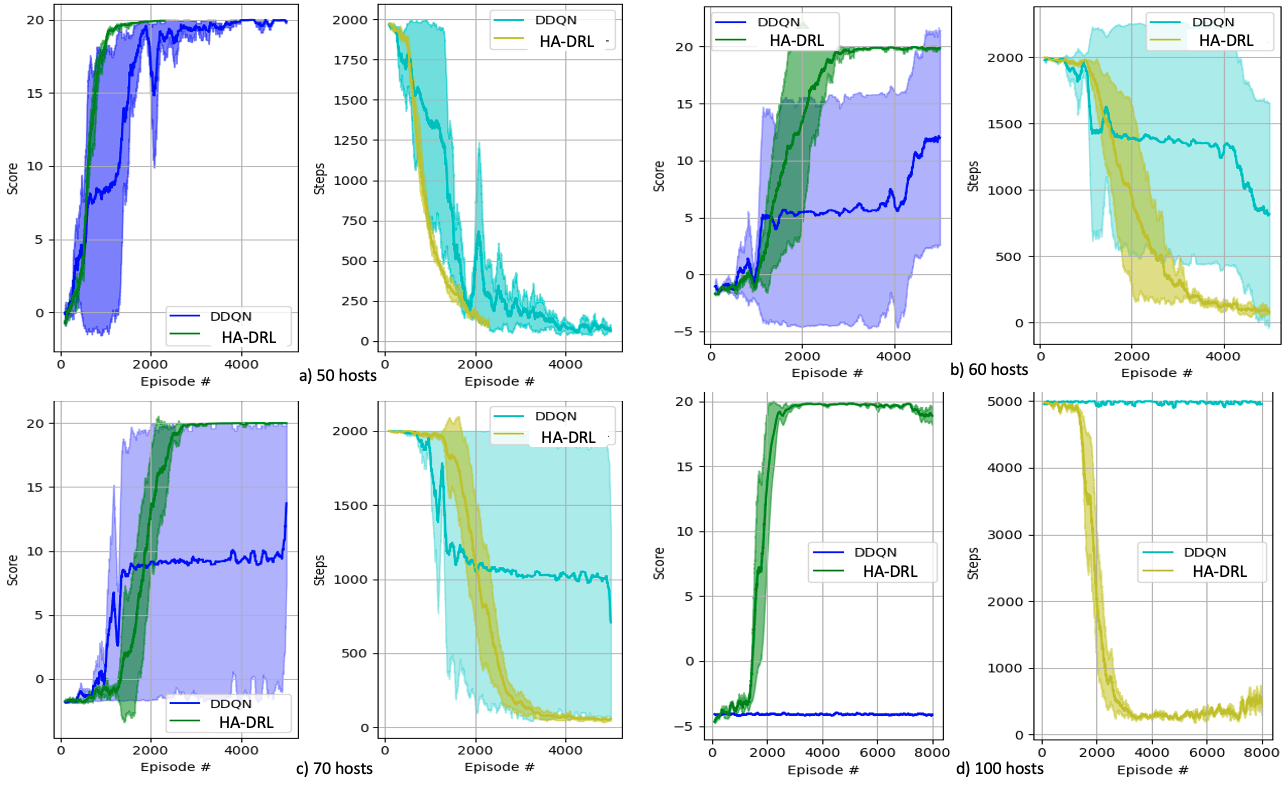}
    \caption{Results of HA-DRL and single DDQN on different CybORG scenarios}
    \label{fig:total_figure}
\end{figure}

Each performance in Figure \ref{fig:total_figure} was repeated five times with different random seeds to ensure reproducibility. The shaded regions on the graphs represent the variability between runs. There are two indicators we are looking at after training the agents: the maximum score the agents can get which is on the left panel and the number of steps to reach the target shown on the right panel. For each scenario, the maximum score the agent can receive is approximately $20$ points (minus some small negative rewards on invalid action it takes to reach the assets).

We have tested the HA-DRL algorithm on a variety of scenarios with different configurations of hosts and action space as shown in Table \ref{table:1}. Regarding scenario complexity, with the number of hosts varied from $6$ to $100$, the action space increases from having $49$ to $4646$ actions. However there is only an addition of having $2$ more agents to train. In all the tested scenarios, HA-DRL convergence results were either similar to or superior to the DDQN agent, depending on the complexity of the scenario and the size of action space.

Figure \ref{fig:total_figure} presents the algorithms' performance on four notable scenarios where the complexity is significant enough to showcase the superiority of the algorithm while not too computationally expensive for repeated training. In scenarios where both DDQN and HA-DRL were able to learn the optimal policy, HA-DRL showed faster and stabler convergence than DDQN. DDQN's performance on scenarios with 60 and 70 hosts were unstable as it only successfully learned the optimal policy in 1 out of 4 runs. HA-DRL held up its performance on the scenario of $100$ hosts where DDQN failed to achieve any progress during training. 

For the $100$ hosts scenario, DDQN was not able to explore the action space and learn the policy at all while HA-DRL took approximately $4000$ episodes to pick up the learning and start to converge to the optimal policy. The policy learned by HA-DRL in each of the tested scenarios is optimal in terms of having minimum number of taken actions.

\begin{table}[ht]
\centering
\begin{tabular}{||c c c c||} 
 \hline
 Hosts & State space & Action space & Agents  \\ [0.5ex] 
 \hline\hline
 6  & 39 & 49 & 2 \\ 
 9  & 55 & 120 & 2 \\
 12  & 71 & 182 & 2 \\
 18  & 217 & 342 & 2 \\
 24  & 285 & 550 & 2 \\ 
 50  & 573 & 1326 & 3 \\
 60  & 685 & 1830 & 4 \\
 70  & 797 & 2414 & 4 \\ 
 100  & 1133 & 4646 & 4 \\ [1ex]
 \hline
\end{tabular}
\caption{Configurations of tested scenarios}
\label{table:1}
\end{table}

This is also the first demonstration of applying deep reinforcement learning in an automatic penetration testing scenario that can handle such a large action space ~\cite{ghanem_chen_2020}.

\subsection{Discussion}
\begin{figure}[ht]
    \centering
    \includegraphics[scale=0.21]{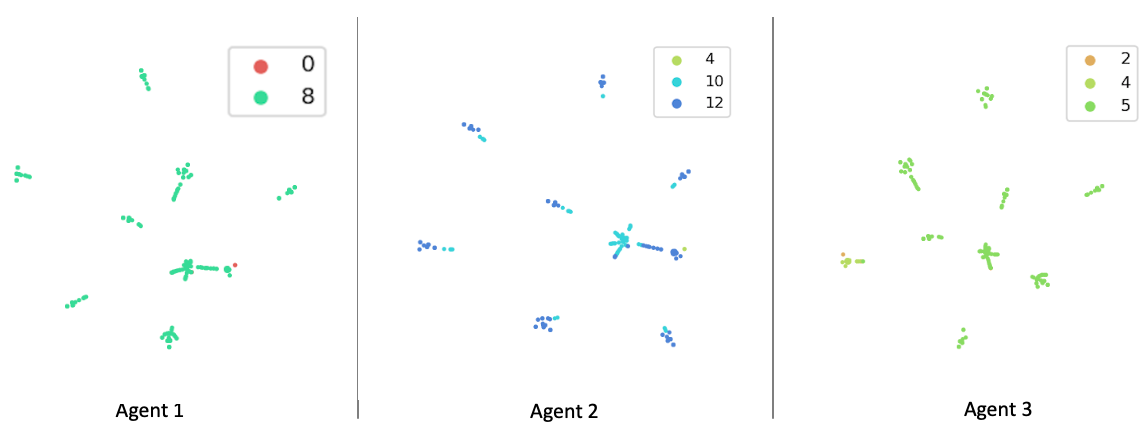}
    \caption{State representation of 50-hosts scenario}
    \label{fig:50_hosts}
\end{figure}

To examine the learning of each agent in HA-DRL, we look into the state and action representation that the trained agents have learned and visualise the state representation using t-SNE method ~\cite{van2008visualizing}. Figure \ref{fig:50_hosts} shows the state representations the agents have learned which are quite similar across the three agents in the architecture used for the 50-hosts scenario. Surprisingly, the learned state representations are mapped into $9$ separate clusters which match the $9$ private subnets in the 50-hosts configuration, which itself has $1$ public subnet and $9$ private subnets each with $5$ hosts, even though this knowledge is not presented or observable to the agents.
The colored marker represents the identifier of the action each agent would take in certain states. The action space of $1326$ is mapped into smaller subsets of $10$ to $15$ actions per agent. After training, each agent learns that only $2$ to $3$ actions in its own subsets are needed to optimally capture the hidden assets. This visualisation opens up new research direction where further look into state and action representation can yield better progress in applying DRL to even more complex PT scenarios.

\section{Conclusions}
This paper introduced a new hierarchical agent reinforcement learning architecture called HA-DRL to tackle PT scenarios with large discrete action space. The proposed algorithm requires minimum prior knowledge on the problem domain while providing competitively better performance than conventional DQN agents. We validate the algorithms on simulated scenarios from CybORG. In all tested scenarios, HA-DRL showed superior performance to the single DDQN agent. 

The proposed algorithm is also scalable into larger action spaces with sub-linear increase in network computational complexity. This is because the individual networks are trained and used for inference separately, and their outcomes are used sequentially to compute the final action. Despite HA-DRL’s promising results, various challenges remain to be resolved. Future works will target extending the HA-DRL into incorporating subgoals learning in order to perform better in environments where rewards are much sparser. Additionally efficient exploration in large action spaces is still a standing problem that deserves further research. 
\section*{Acknowledgements}
This work was supported by the Australian Defence Science Technology Group (DSTG) under Agreement No: MyIP 10699.

\appendix

\bibliographystyle{named}
\bibliography{ijcai21}

\end{document}